\documentclass[sn-apa]{sn-jnl}
\usepackage{graphicx}%
\usepackage{amsmath,amssymb,amsfonts}%
\usepackage{amsthm}%
\usepackage{mathrsfs}%
\usepackage[title]{appendix}%
\usepackage{textcomp}%
\usepackage{manyfoot}%
\usepackage{booktabs}%
\usepackage{algorithm}%
\usepackage{algorithmicx}%
\usepackage{algpseudocode}%
\usepackage{listings}%
\usepackage{comment}
 \usepackage[table,xcdraw]{xcolor}
\usepackage{subfig}
\usepackage{multirow}
\usepackage{xspace} 
\usepackage{rotating}
\usepackage{tikz}
\usepackage{xcolor,colortbl}
\usepackage{natbib}
\setcitestyle{authoryear,open={(},close={)}}

\theoremstyle{thmstyleone}%
\theoremstyle{thmstyletwo}%

\theoremstyle{thmstylethree}%

\usepackage{pdfpages}
\begin{document}

\title[Article Title]{Can Explainable AI Assess Personalized Health Risks from Indoor Air Pollution?}








\author*[]{\fnm{*Pritisha} \sur{Sarkar}}\email{ps.19cs1110@phd.nitdgp.ac.in}

 \author[]{\fnm{Kushalava reddy} \sur{Jala}}\email{krj.20u10735@btech.nitdgp.ac.in}



\author[]{\fnm{Mousumi} \sur{Saha}}\email{mousumi.saha@cse.nitdgp.ac.in }


\affil[]{\orgdiv{Department of Computer Science and Engineering}, \orgname{National Institute of Technology, Durgapur}, \orgaddress{\street{M.G.Avenue}, \city{Durgapur}, \postcode{713209}, \state{West Bengal}, \country{India}}}


\abstract{Acknowledging the effects of outdoor air pollution, the literature inadequately addresses indoor air pollution's impacts. Despite daily health risks, existing research primarily focused on monitoring, lacking accuracy in pinpointing indoor pollution sources. In our research work, we thoroughly investigated the influence of indoor activities on pollution levels. A survey of 143 participants revealed limited awareness of indoor air pollution. Leveraging 65 days of diverse data encompassing activities like incense stick usage, indoor smoking, inadequately ventilated cooking, excessive AC usage, and accidental paper burning, we developed a comprehensive monitoring system. We identify pollutant sources and effects with high precision through clustering analysis and interpretability models (LIME and SHAP). Our method integrates Decision Trees, Random Forest, Naive Bayes, and SVM models, excelling at 99.8\% accuracy with Decision Trees. Continuous 24-hour data allows personalized assessments for targeted pollution reduction strategies, achieving 91\% accuracy in predicting activities and pollution exposure.
}




\maketitle



\section{Introduction}
Invisible yet impactful, air pollution silently weaves its way into the fabric of our environment, leaving behind a trail of consequences.
As the air we breathe carries both life and danger, the insidious effects of air pollution on human health have become an undeniable reality. Air pollution refers to the presence of harmful substances in the air, stemming from various sources like industry, transportation, and natural activities. Outdoor air pollution pertains to the contamination of the external atmosphere by pollutants, while indoor pollution involves the presence of harmful substances within enclosed spaces, often resulting from activities like cooking \cite{b37}, smoking \cite{b38}, and the use of certain products.
Even indoor sources like air coolers and incense sticks emit various pollutants, posing health risks. The health impacts of these pollutants are substantial, with short-term exposure leading to coughing, wheezing, and breathing difficulties and long-term exposure causing cardiovascular and respiratory diseases.
Volatile Organic Compounds (VOCs) \footnote{\url{https://rb.gy/hwetd}}, nitrogen dioxide (NO$_{2}$) \footnote{\url{https://www.epa.gov/no2-pollution/basic-information-about-no2}}, PM$_{10}$ \footnote{\url{https://pubmed.ncbi.nlm.nih.gov/15098563/}}, and PM$_{2.5}$ are released from activities like cooking, smoking, burning, and air conditioner use, adversely affecting human health. While outdoor air pollution's health impacts are widely recognized, indoor air pollution awareness remains alarmingly low. 


An online survey with 143 participants highlighted that a significant portion, around 51\%, fell within the 25 to 35 age group, and 25.9\% are aged 18-25. Despite spending time indoors and outdoors, only 3\% of respondents are aware of indoor activities' impact on air quality. Smoking was identified as a pollutant source by 37\% of participants, while only 6\% recognized cooking without proper ventilation as problematic. Vehicles and construction are recognized by approximate 90\% . 


Several research papers have focused on detecting indoor pollution risk scenarios. However, existing works primarily concentrate on indoor air quality monitoring and alert systems \cite{b22,b26,b34} rather than pinpointing which indoor activities cause pollution and health effects. For instance,
\cite{b5} developed a system to detect house fires early, while \cite{b17} designed a smart wearable badge to monitor pollutants related to personal activities. Additionally, \cite{b29} created a system to identify elderly activity, and \cite{b3} presented a low-cost indoor air quality monitoring device that measures CO, NO$_{2}$, and PM$_{2.5}$ levels. The research work by \cite{b30} emphasizes the importance of accurate estimates of human exposure to air pollutants for risk appraisal and control strategies.  
Our research seeks to address the gap by not only monitoring but also identifying the key indoor activities responsible for pollution. By collecting individuals' specific data on indoor activities and pollutant levels, we aim to shed light on the primary contributor to indoor air pollution with their corresponding activity and empower informed decision-making.

We concentrate on prevalent indoor activities such as incense stick usage, indoor smoking, poorly ventilated cooking, and accidental paper burning. By identifying and predicting the effects of pollutant sources on human health, we strive to offer effective solutions. Our ultimate goal is to arm individuals with a comprehensive understanding of their exposure for indoor activity, moving beyond just measurement to facilitate well-informed actions. We aspire to bridge the gap between awareness and action, ensuring individuals to make informed decisions that mitigate the harmful effects of indoor air pollution by their activity.

\subsection{Problem statement}
 Can we start by considering our daily routines and behaviours after identifying instances where personal activities may contribute to air pollution? Can we investigate the ability of our model to identify ambient situations and estimate whole-day human exposure to air pollutants?

\subsection{Contribution}
The contribution of this work can be summarized as follows:
\begin{itemize}
    \item \textbf{Comprehensive data collection:} The study collects data on various pollutant sources in different situations: ideal and worst. This allows for a more comprehensive understanding of the different scenarios that contribute to pollution levels. Prior to analysis, the dataset undergoes pre-processing, which may include cleaning, normalization, and feature extraction. This step ensures that the data is in a suitable format for further analysis.
    \item \textbf{Identifying key features for indoor pollution:} Using clustering analysis we group similar situation-based pollutants, revealing patterns and pollution sources. LIME \cite{lime} and SHAP \cite{shap} models interpret the significance of each cluster, offering insights into contributors to pollution benchmark exceedance, using interpretable explanations for complex machine learning predictions on the clustered data.
    \item \textbf{Individualized assessment:} By collecting 24 hours of data on a person's daily pollutant exposure, the study aims to identify the main reason or situation responsible for exceeding the pollution benchmark. This individualized assessment helps to understand the specific factors contributing to high pollution levels for that person, enabling targeted interventions or lifestyle changes by predicting human activity.
    \end{itemize}
Overall, our work combines data collection, clustering analysis, and interpretability techniques to gain insights into the contribution of different situations and sources to pollution levels. It provides a personalized pollution assessment and management framework, enabling targeted interventions and informed decision-making to improve air quality and protect individual health.

\subsection{Article organization}
The paper is structured as follows: 
Section \ref{sec2} presents a thorough discussion of related works, focusing on air quality sensing and indoor air quality sensing and the development of some models for various research purposes.
Section \ref{sec3} depicts outlines the workflow of our proposed research. In Section \ref{sec4}, we present a comprehensive review of our targeted pollutants, accompanied by data visualization through plots to illustrate the collected data. Section \ref{sec5} delves into the methodology, while Section \ref{1} and \ref{2} focuses on evaluating our approach. Finally, we conclude our work in Section \ref{sec7}.

\section{Literature survey}\label{sec2}
Researchers have conducted extensive studies on indoor air quality sensing and exposure estimation. The studies are categorized into three sections: personalized devices for real-time indoor air quality information, alert systems to address indoor activity pollution and understanding of health impacts. This comprehensive overview aims to raise awareness and promote healthier indoor environments for individuals. 

\subsection{Health implications for indoor air quality}
In research \cite{b30}, Ken Sexton and P. Barry Ryan emphasized accurate estimates of human exposure to inhaled air pollutants to appraise risks and develop mitigation strategies. They focused on outdoor and indoor environments, utilizing Statistical, Physical, and Physical-Stochastic Modelling to assess the concentration, Exposure, and Dose of pollutants. The study proposed evaluating the indirect approach, identifying crucial micro-environments, and complementing direct exposure measurements in future work. 



In another work \cite{b31} Kerri A. Johannson et al. provides compelling evidence of the association between ambient air pollution and interstitial lung disease (ILD). They explore mechanisms such as oxidative stress and inflammation, linking air pollution exposure to ILD development. The study introduces the "exposome" concept to understand pulmonary fibrosis pathobiology, advocating for comprehensive research on environmental contributors to ILD and potential improvements in patient outcomes.
In the PURE-AIR study \cite{b32}, Michael Jerrett and Papreen Nahar Siddiquea conducted a randomized controlled trial across 120 rural communities in eight countries to investigate the health and climate impacts of clean cooking fuels. 
The research work \cite{b36} presented in this paper contributes by proposing a novel architecture for automatically recognizing food preparation activities at home using air quality data from commercial sensors. This approach addresses the challenges faced by frail individuals, especially the elderly, who have difficulty maintaining food diaries consistently. Table \ref{STUDY} provides a literature survey of various papers focused on \textit{Indoor Air Quality sensing and alert systems for prevent worst situation}.


\begin{table}[]
\caption{Literature survey--Indoor air quality sensing and alert system}\label{STUDY}
\tiny
\begin{tabular}{|l|l|l|l|l|l|}
\hline
\textbf{Ref.} &
  \textbf{Features} &
  \textbf{Contribution} &
  \textbf{Scenario} &
  \textbf{Result} &
  \textbf{Future scope} \\ \hline
\cite{b22} &
  \begin{tabular}[c]{@{}l@{}}LPG, Alcohol, \\ Hydrogen, CO,\\ temperature\end{tabular} &
  \begin{tabular}[c]{@{}l@{}}Intelligent kitchen \\ safety monitor, \\ automatic gas/power \\ cutoff and fire\\  response\end{tabular} &
  Cooking &
  \begin{tabular}[c]{@{}l@{}}Significant \\ electricity \\ cost reduction\end{tabular} &
  \begin{tabular}[c]{@{}l@{}}Enhance \\ Home Device \\ Prediction\end{tabular} \\ \hline
\cite{b26} &
  \begin{tabular}[c]{@{}l@{}}Temperature, \\CO, humidity,\\VOC\end{tabular} &
  \begin{tabular}[c]{@{}l@{}}Detection of risk\\ situations around \\ oven\end{tabular} &
  Cooking &
  \begin{tabular}[c]{@{}l@{}}Efficiently detects \\ and  manages risk \\ situations\end{tabular} &
  \begin{tabular}[c]{@{}l@{}}Evaluate engine \\ in real-world \\ cooking\end{tabular} \\ \hline
\cite{b34} &
  VOC, CO$_{2}$ &
  \begin{tabular}[c]{@{}l@{}}Detection of \\ indoor smoking\end{tabular} &
  Smoking &
  \begin{tabular}[c]{@{}l@{}}Using IoT sensors, \\ achieve 93\%\\ accuracy\end{tabular} &
  \begin{tabular}[c]{@{}l@{}}Optimize feature \\ extraction\end{tabular} \\ \hline
\cite{b15} &
  \begin{tabular}[c]{@{}l@{}}PM$_{2.5}$, VOC\end{tabular} &
  \begin{tabular}[c]{@{}l@{}}Automatic\\ detection of \\ pollution events\end{tabular} &
  \begin{tabular}[c]{@{}l@{}}Cooking, \\ smoking, \\ spraying \\ pesticide\end{tabular} &
  \begin{tabular}[c]{@{}l@{}}Successfully \\ identifies pollution \\ sources\end{tabular} &
  \begin{tabular}[c]{@{}l@{}}Forecast future\\  air quality\end{tabular} \\ \hline
\cite{b17} &
  \begin{tabular}[c]{@{}l@{}}CO$_{2}$ , \\ VOCs\end{tabular} &
  \begin{tabular}[c]{@{}l@{}}Design smart \\ cooking wearable\\  badge\end{tabular} &
  Cooking &
  \begin{tabular}[c]{@{}l@{}}Accuracy achieved \\ 92.44\%\end{tabular} &
  \begin{tabular}[c]{@{}l@{}}Optimizing \\ Hardware\\ Platform\end{tabular} \\ \hline
\cite{b16} &
  VOCs, PM &
  \begin{tabular}[c]{@{}l@{}}Air quality \\ socializing\end{tabular} &
  -------- &
  \begin{tabular}[c]{@{}l@{}}Air Quality \\  Awareness\end{tabular} &
  \begin{tabular}[c]{@{}l@{}}Improving Air  \\ quality \\ Information\end{tabular} \\ \hline
\cite{b5} &
  \begin{tabular}[c]{@{}l@{}}Smoke, gas,  \\ temperature\end{tabular} &
  \begin{tabular}[c]{@{}l@{}}Early detection\\ for  house fires\end{tabular} &
  ---------- &
  \begin{tabular}[c]{@{}l@{}}System detects early \\ fire and conserves \\ energy\end{tabular} &
  \begin{tabular}[c]{@{}l@{}}Employ mobile   \\ monitoring \\ equipment\end{tabular} \\ \hline
\cite{b21} &
  \begin{tabular}[c]{@{}l@{}}PM$_{2.5}$, \\ PM$_{10}$\end{tabular} &
  \begin{tabular}[c]{@{}l@{}}Feasible monitoring \\ system\end{tabular} &
  ------ &
  \begin{tabular}[c]{@{}l@{}}78\% - 95\% Indoor \\ and Outdoor testing\end{tabular} &
  \begin{tabular}[c]{@{}l@{}}Flexible\\ Miniaturization\end{tabular} \\ \hline
\end{tabular}
\end{table}

\subsection{Indoor air quality sensing}
In some previous research studies have shown that polluted air uses IoT sensors and instruments to demonstrate what happens in the air. The sensors are used for the monitoring of gases in the air. They targeted on pollutants like CO, O$_{3}$ and NO$_{2}$. They employed the Arduino model for their monitoring system. The authors also discussed their future work, which includes launching a three-phase air pollution surveillance system aimed at monitoring global air emissions. Some study \cite{b19} has found an association between exposure to traffic-related air pollution. The study \cite{b3} critically summarizes recent research on the development of indoor air quality monitoring devices using low-cost sensors, highlighting a lack of studies with adequate calibration and validation, suggesting a need for more research in this area to ensure data reliability and standardization.
Some work \cite{b13,b11,b12,b20,b18} the researchers develop a system that estimates personal air pollution inhalation dosage by combining data from a mobile app interfaced with wearable personal activity sensors for breathing rate and participatory pollution monitoring, conducting field trials in Sydney to demonstrate the system's ability to provide real-time pollution inhalation dosage estimates based on different levels of activity, leading to more accurate medical inferences compared to earlier systems without personal activity information.
Some study \cite{b35,b28} contributes to the understanding of the potential impact of global climate change, demographic shifts, and advancing mechanization on living conditions in central Europe. They highlight the evolving concept of smart homes, which now encompasses energy efficiency, home automation, communication networks and also wearable devices for environmental monitoring. The review presents insights into the effect of environmental parameters on indoor air quality, thermal comfort, and living behavior in smart homes, as well as evaluating sensor technologies, data security aspects, and the acceptance of smart technologies by occupants. In research study \cite{b6} introduce a vehicular-based approach for real-time fine-grained air quality measurement using cost-effective data farming models, including one deployed on public transportation and another as a personal sensing model mounted inside a vehicle to measure carbon monoxide levels that correlate with outdoor values. The study \cite{b8} was to design and integrate Internet of Things (IoT) technology into an Arduino-based fire safety system, enabling real-time alerts to fire-fighting facilities, authorities, and building occupants to prevent fire occurrences or minimize potential damages and improve overall safety in the face of increasing fire incidents.

\section{System overview}\label{sec3}
\begin{figure}
    \centering
    \includegraphics[width=7cm]{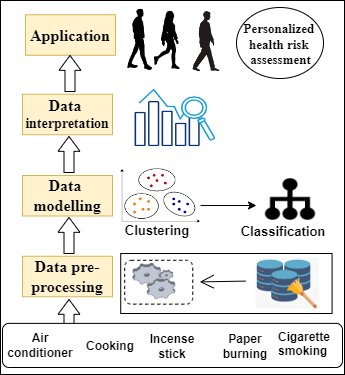}
    \caption{System framework for indoor exposure, health risk assessment, activity prediction, with data collection, analysis, visualization, interpretation of our research work}\label{flow1}
\end{figure}

{In our study, we sought to develop a methodology for personalized health assessment by collecting indoor pollutant data from various sources using the FLOW air quality monitoring device Figure (\ref{situation}) A.
Figure \ref{flow1} provides a broad overview of the framework proposed to answer our problem statement. 
The pollutants of interest were VOC, NO$_{2}$,  PM$_{10}$, PM$_{2.5}$, and PM$_{1}$, which are known to be particularly harmful to human health and living organisms. We undertook essential data pre-processing steps to ensure accurate and meaningful analysis, such as handling null values, normalization, and removing negative values to focus solely on relevant and positive data points, thereby enhancing the accuracy of our subsequent analysis.
We employed clustering analysis to group pollutants based on similar situations in the data modelling stage. Additionally, we utilized machine learning algorithms to validate the accuracy of our findings.
We employed LIME (Local Interpretable Model-Agnostic Explanations) and SHAP (Shapley Additive exPlanations) models for data interpretation. LIME provided interpretable explanations for predictions made by complex machine learning models, enabling us to gain insights into which situations or pollutant sources contributed the most to pollution exceeding the benchmark levels. 
The application section focused on Personalized Health Risk Assessment concerning indoor air pollution. Understanding the specific situations or sources responsible for elevated pollution levels in individuals allowed us to conduct personalized health risk assessments. With knowledge of the impact of different conditions on an individual's pollutant exposure, healthcare professionals can provide tailored recommendations to mitigate the adverse effects of pollution on their health.
We've devised a robust method for personalized health assessment by analyzing indoor pollutant data, aiding targeted pollution reduction strategies. Current research lacks a comprehensive assessment of pollution from mixed indoor scenarios, crucial for understanding daily exposure. Our goal is a model that gauges cumulative indoor pollution intake over 24 hours, vital for improving health and air quality..}

\section{Data description}\label{sec4}
\begin{figure*}
    \centering
    \includegraphics[width=\textwidth]{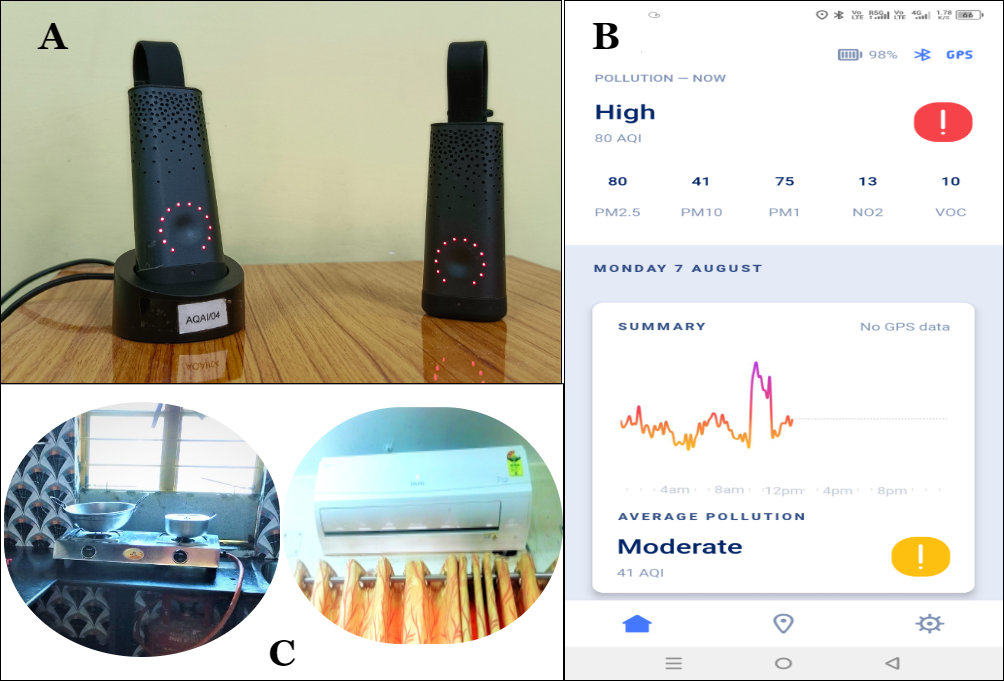}
    \caption{A. Illustration depicting the interface of a Flow device and B. showcasing its components and functionalities. C. Instances of Data Acquisition Scenarios, like cooking, AC usage.}\label{situation}
\end{figure*}

In our research work, we aimed to devise a personalized health assessment methodology by utilizing the FLOW air quality monitoring device (in Figure \ref{situation} A.) to gather indoor pollutant data from diverse sources. 
We conducted one-month data collection for each scenario, with ventilation present during incense stick burning and smoking but no ventilation in other situations. The data was collected in closed rooms of similar sizes using FLOW air quality monitoring devices. Indoor air pollution sources like gas oven cooking, wooden stove cooking, paper burning, smoking, air conditioning, and incense stick burning emit harmful pollutants such as nitrogen dioxide, carbon monoxide, particulate matter, and volatile organic compounds, which can negatively impact indoor air quality and human health, leading to respiratory problems, headaches, asthma, and other health issues. Implementing proper ventilation and utilizing air purification systems can be effective in mitigating the effects of these pollutants.



\subsection{Data distribution}
To collect indoor air pollution data in various indoor settings (Figure \ref{situation} C.), we employed the Plume Labs' Four FLOW (version 2) apparatus. This device captures raw air data and assesses concentrations of VOC, NO$_{2}$, PM$_{10}$, PM$_{2.5}$, and PM$_{1}$ , the interface illustrated in Figure \ref{situation} B. In another laboratory study \cite{lab}, we verify the data generated by this device. In the cooking data  VOC stands out with an exceptionally high mean, upper limit, and a wider range of data compared to any other source or pollutant. The only other pollutant with data exceeding 200 µg/$m^3$ is PM$_{10}$.
In the AC plot, VOC also exhibits a high mean, upper limit, and a wider data range than any other pollutant, but unlike in cooking, there are no outliers present. Similar to the previous plot, Here, PM$_{10}$ is the only other pollutant with data reaching near 200 µg/$m^3$ .
In the paper burning plot, VOC shows a high variance but a smaller range of about 50 when compared to AC and Cooking, both of which have a much larger range of almost 300. The upper limit for VOC is lower than the other sources but still higher than the other pollutants. PM$_{10}$, like the other sources, has an upper limit of less than 50 and does not have data near or exceeding 200 µg/$m^3$.
In the incense stick plot, VOC data is split into two parts: one ranging from 300 ppb to 400 ppb, which is more common with AC, and the other ranging from 100 to 250, similar to paper burning. PM$_{10}$ data is almost as much as VOC, and it has a wider range than VOC or any other pollutant. Unlike previous sources, it has a significant amount of data above 200.
In the smoke plot, VOC again exhibits high variance but a smaller range of 200 ppb, which is the second lowest, only behind paper burning.\\
Overall, VOC is a major concern across all pollution sources, with cooking being the primary contributor. Understanding the characteristics of different pollutants from various sources is crucial for devising targeted pollution control strategies and promoting cleaner air quality.

\section{Methodology}\label{sec5}
Our primary goal is to create a learning model that identifies personal exposure to air pollution from various activities, which helps to develop effective pollution control strategies. The model will recommend appropriate precautions to mitigate health and environmental impacts caused by indoor activities in the air. Steps to assess the individual’s pollution intake;

\begin{itemize}
    \item Identify the major pollutants from different indoor pollutant activities.
    \item Individuals’ indoor pollution exposure calculation.
\end{itemize}

\textbf{Features and its significance -}
In our study, we focused on investigating five distinct indoor pollutant sources, namely \textbf{'AC (air conditioning)'}, \textbf{'Paper burn'}, \textbf{'Incense stick usage'}, \textbf{'Cooking'}, and \textbf{'Cigarette smoking'}. For each of these sources, we monitored and analyzed five specific pollutants: f$_{no2}$, f$_voc$, f$_{pm10}$, f$_{pm2.5}$, and f$_{pm1}$. By categorizing the sources as S(X) = {X$_1$, X$_2$, X$_3$, X$_4$, X$_5$} and the pollutants as S(X) = {f$_{no2}$, f$_{voc}$, f$_{pm10}$, f$_{pm2.5}$, f$_{pm1}$}, we were able to conduct a comprehensive assessment of the indoor air quality and its potential impact on human health and well-being.

\subsection{Identify the major pollutants from different indoor pollutant activities}
Our approach to identifying the major pollutants in indoor pollution activities involves several steps. First, we apply \textbf{\textit{"Clustering methods"}} to group pollutants based on similarities. Using these clusters, we then employ \textbf{\textit{"Machine learning algorithms}} to predict the likelihood of each pollutant being the major one in the dataset. Next, we utilize \textbf{\textit{"Interpretable models"}} to understand the weightage of pollutants and determine the final outcome, thereby identifying the major pollutants accurately. This process helps us pinpoint and prioritize the most significant pollutants in indoor pollution scenarios.

\subsubsection{Clustering of indoor air pollutants}
Clustering is an unsupervised technique that organizes a dataset by identifying hidden patterns and grouping data into distinct clusters based on their similarities. This approach enabled us to identify patterns and commonalities among pollutants, shedding light on potential sources of pollution.

\subsubsection{Data modeling}
By utilizing cluster features, we've built a multidimensional feature space, which is subsequently investigated using diverse machine learning models. The goal is to decipher the semantic context of an unfamiliar dataset related to a particular region. We extensively train and test acquired clusters using various machine learning models to pinpoint pertinent data points within each cluster, ultimately aiming to find a model accurately encapsulating cluster-specific characteristics. In pursuit of this goal, we conducted experiments using a variety of machine learning models to evaluate their effectiveness and suitability for the task.
\begin{itemize}
    \item \textbf{Decision tree(DT) -}   Decision tree \cite{tree} is a tree-like structure that classifies data by sorting based on feature values. Internal nodes represent data features, branches depict feature values, and leaf nodes represent the final model output. 

    \item  \textbf{Naive Bayes(NB) -}  In indoor air pollution, Naive Bayes \cite{nb} is utilized as a probabilistic classifier to estimate the likelihood of specific pollutants being present based on observed features or sensor readings. It assumes independence between features, allowing for efficient and straightforward calculations even with limited data, making it a popular choice for real-time indoor air quality monitoring systems.

    \item  \textbf{Random forest algorithm(RF) -} It is an ensemble of numerous decision trees, each providing a class prediction \cite{rand}. The model's final prediction is determined by aggregating the individual tree's predictions, with the class receiving the most votes becoming the chosen outcome.

    \item  \textbf{Support vector machine(SVM) -}  Support Vector Machines (SVMs) \cite{svm} are employed in indoor air pollutant prediction as a powerful machine learning algorithm to classify and predict pollutant levels based on sensor data. SVMs aim to find an optimal hyperplane that best separates the data into different pollutant categories, allowing for accurate and robust predictions in complex indoor environments.

\end{itemize}

\subsubsection{Data interpretation} In our study, we utilized the LIME (Local Interpretable Model-Agnostic Explanations) framework and SHAP(SHapley Additive exPlanations model, an open-sourced tool, to understand the dominant features and their significance within each cluster.
\begin{itemize}
    \item \textbf{LIME Model -} LIME enables us to explain the behavior of any black-box classifier in a reliable manner and assess the validity of the classifier's analysis locally. It works by taking a learning model as input, making it applicable to any machine learning model. LIME provides easily understandable explanations by analyzing the model's behavior based on input data samples and explaining how predictions change with internal characteristics. We used Random Forest for implementing LIME. The results and inferences from the LIME model are discussed in the subsequent section after a successful training and testing process, and selecting the most suitable model.\\
\item \textbf{SHAP model -} Machine learning models are becoming highly accurate yet complex "black boxes." To address this, the focus on interpretability and explainability is increasing. SHAP values are a promising tool, measuring feature contributions (e.g., income, age, credit score) to predictions, providing transparency and understanding of the model's decision-making process.
\end{itemize}

\begin{algorithm}
\textbf{Algorithm 1 -
Clusters potency :}
Using the pollutants' weight factors, the subsequent section will rank the evaluated clusters by following the steps outlined below. 

\textbf{Input:}
\begin{itemize}
    \item  A set of m clusters, denoted as \{$C_0$,$C_1$, ..., $C_m$\}.
    \item Each cluster $C_i$ contains n number of pollutant data $P_{j}$, denoted as \{$P_1$, $P_2$, ..., $P_n$\}
  
    \item Each pollutant data $P_{jk}$ has k instance, denoted as \{$P_{j1}$, $P_{j2}$, ..., $P_{jk}$\}.
    \item Weight factors for pollutants, denoted as \{$W_1$, $W_2$, ..., $W_k$\}.
\end{itemize}

\textbf{Output:}
- A ranked list of clusters based on the mean value of pollutants.\\
\textbf{Steps:}\\
\textbf{Step 1.} For each cluster $C_i$ , where i = \{0,1,2,...,m\}

~~~~~~~~~~~For each pollutant data $P_j$ in $C_i$ , where j = \{1,2,...,n\}

~~~~~~~~~~~For each pollutant data $P_j$ has k no. of  instance $P_{jk}$, where k = \{1,2,...,k\}

~~~~~~~~~~~Multiply the value of $P_{jk}$ by its corresponding weight factor $W_k$\\
\textbf{Step 2.} Calculate the sum of all the weighted pollutant values in $C_i$ for pollutant 

~~~~~~~~~~~$P_{jk}$, denoted as Sum($C_i$($P_{jk}$)).

~~~~~~~~~~~Calculate the mean value of pollutant $P_{jk}$ in $C_i$, denoted as, Mean($C_i$($P_{jk}$)),



\textbf{Step 3.} Compare the mean values of pollutants for each cluster $C_i$ , where i = \{1,2,...,m\}.\\
\textbf{Step 4.} Identify the cluster $C_m$ that has the highest mean value of pollutants

~~~~~~~~~~ among all clusters:

~~~~~~~~~~ So, $C_m$ = argmax(Mean($C_i$($P_{jk}$)), where i = \{1,2,...,m\}, and k = \{1,2,...,k\}.
\textbf{Step 5.} Rank the clusters based on their mean pollutant values:

~~~~~~~~~~ Cluster 1: $C_m$ (The cluster with the highest mean value of pollutants).

~~~~~~~~~~ Cluster 2: Next cluster with the second-highest mean value of pollutants.

~~~~~~~~~~~~~~Cluster 3: The next cluster with the third-highest mean value of pollutants.

~~~~~~~~~~~~- ... and so on, until all clusters are ranked.\\
\textbf{Step 6.} Output the ranked list of clusters based on their mean pollutant
values, named as Cluster potency $(C_x)$.
\end{algorithm}

After completing the clustering and modelling processes, employing the LIME and SHAP models aids in understanding the significance of indoor air pollutants of each cluster. Consequently, we can determine the weight factors of each indoor pollutant based on their respective importance levels. This approach allows us to accurately assess the impact of individual indoor air pollutants and prioritize them accordingly.

\begin{figure}[htbp]
    \centering
    \includegraphics[width=12cm]{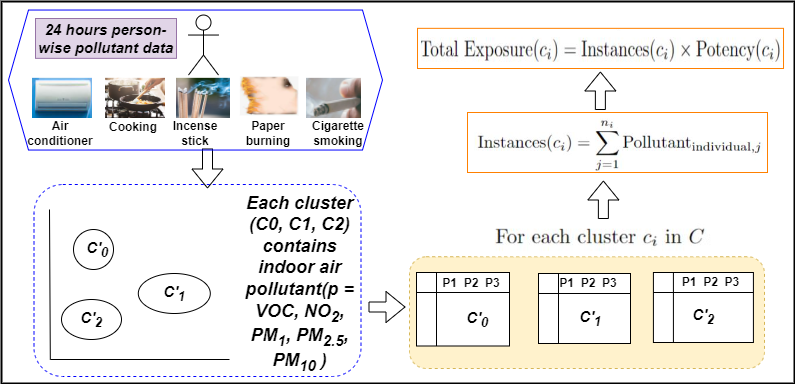}
    \caption{Workflow to calculate individual's indoor air pollutant intake}\label{indoor}
\end{figure}

\subsection{Individuals’ personalize indoor air pollution exposure calculation}

To assess an individual's personal pollution intake, we start by determining the pollutants cluster Ranking . This involves calculating the sum and mean of each pollutant within each cluster and finding the ratio of the maximum pollutants in each cluster, defining it as \textbf{\textit{"Cluster potency"}} (in detail explain in Algorithm 1).
Next, we gather 24-hour pollutant data for an individual person and perform clustering again. The number of clusters remains the same as in our previous analysis (e.g., if we initially obtained c=z clusters, the individual's pollutant cluster number will also be c=z). We then calculate the instances of pollutants in each cluster (e.g., 1st cluster has 20 instances, 2nd cluster has 50 instances, and so on).

Finally, we multiply the total instances of each cluster by its cluster potency. This product is termed as 'Total exposure', representing the individual's cumulative pollutant intake within each cluster based on their 24-hour pollutant data. This comprehensive approach helps us gauge and analyze the specific pollutants influencing an individual's pollution intake. The whole procedure explains through the Figure \ref{indoor}.








\section{Evaluation : identify the major pollutants from different indoor pollutant activities }\label{1}
In this section, at first we apply clustering algorithm in our dataset, then we employed multiple classification Machine Learning models, including Random Forest, Decision Tree, Naive Bayes, and SVM, in our proposed approach. Performance metrics such as F1 score, Precision, Recall, and Kappa Statistics are used for evaluation in Section \ref{eva}. After that data is interpreted using LIME and SHAP models in Section \ref{inter}. Our study concludes by highlighting the significance of the outcomes obtained from the comprehensive dataset analysis.\\

\subsection{Clustering of indoor air pollutants}
In this research, we adopt the K-means \cite{kmeans} algorithm due to its efficiency and simplicity in creating clusters from sample points. To determine the optimal number of clusters (k=3), we employ the Elbow Method, which calculates Within-Cluster-Sum of Squared Errors (WCSS) for various k values, selecting the point where WCSS starts diminishing as the ideal k value. We have checked the clusters using elbow method and silhouette score, silhouette score is used to evaluate the goodness of clusters. If the silhouette score value is near to 1, then clustering is good in our clustering we got the silhouette score as 0.84.. Additionally, feature scaling is applied to normalize data within a specific range. Analyzing cluster patterns, both with and without feature scaling, reveals distinct clusters depicting indoor air pollutants through the K-means algorithm. These clusters are then explored to gain deeper insights into the characteristics of the samples they encompass.

\begin{table}[]
\footnotesize
\caption{Summary of Decision tree model's accuracy at various sample sizes}\label{sample}
\begin{tabular}{|l|l|l|l|l|l|l|l|l|l|}
\hline
\textbf{Instance} &
  \textbf{Accuracy} &
  \textbf{\begin{tabular}[c]{@{}l@{}}Kappa \\ statics\end{tabular}} &
  \textbf{\begin{tabular}[c]{@{}l@{}}Root Mean \\ Squared \\ Error\end{tabular}} &
  \textbf{\begin{tabular}[c]{@{}l@{}}Matthews \\ Correlation \\ Coefficient\end{tabular}} &
  \textbf{F1 Score} &
  \textbf{Precision} &
  \textbf{Recall} \\ \hline
1000 & 0.9933 & 0.941  & 0.163 & 0.943 & 0.965 & 0.93  & 0.998 \\ \hline
2000 & 0.9966 & 0.980  & 0.11  & 0.980 & 0.989 & 0.998 & 0.981 \\ \hline
3000 & 0.9944 & 0.962  & 0.14  & 0.96  & 0.97  & 0.97  & 0.98  \\ \hline
4000 & 0.9958 & 0.969  & 0.12  & 0.97  & 0.985 & 0.97  & 0.99  \\ \hline
5000 & 0.9975 & 0.987 & 0.89  & 0.98  & 0.993 & 0.991 & 0.995 \\ \hline
6000 & 0.9982 & 0.985  & 0.047 & 0.985 & 0.993 & 0.999 & 0.988 \\ \hline
6500 & 0.9984 & 0.989  & 0.078 & 0.989 & 0.995 & 0.994 & 0.997 \\ \hline
7000 & 0.9985 & 0.9908 & 0.002 & 0.990 & 0.955 & 0.999 & 0.992 \\ \hline
7500 & 0.9986 & 0.985  & 0.094 & 0.985 & 0.992 & 0.996 & 0.989 \\ \hline
8000 & 0.9987 & 0.992 & 0.07  & 0.992 & 0.996 & 0.999 & 0.993 \\ \hline
8100 & 0.9995 & 0.9973 & 0.040 & 0.997 & 0.998 & 0.999 & 0.998 \\ \hline
8200 & 0.9991 & 0.9949 & 0.05  & 0.994 & 0.997 & 0.997 & 0.998 \\ \hline
8300 & 0.9992 & 0.9864 & 0.089 & 0.986 & 0.993 & 0.999 & 0.988 \\ \hline
\rowcolor{yellow}
\textbf{8500} & \textbf{0.9996} & \textbf{0.997}  & \textbf{0.03}  & \textbf{0.997} & \textbf{0.998} & \textbf{0.997} & \textbf{1.00}  \\ \hline
{8600} & 0.9984 & 0.9895 & 0.078 & 0.989 & 0.994 & 0.994 & 0.995 \\ \hline
{8700} & 0.9984 & 0.9905 & 0.07  & 0.99  & 0.995 & 0.997 & 0.994 \\ \hline
\end{tabular}
\end{table}

\subsection{Data modeling : Machine learning model evaluation}\label{eva}
In our model evaluation, we utilized a dataset consisting of 10,559 instances for training and testing purposes. The dataset is split into 70\% for training and 30\% for testing. It is evident that the Decision Tree model achieved the highest F1 score among all other models. SVM and Random Forest displayed similar F1 scores, while Naive Bayes showed a poor F1 score. Moreover, the Decision Tree model exhibited the highest Matthews correlation, whereas Naive Bayes had the lowest correlation. In terms of precision, Random Forest, SVM, and Decision Tree models showed almost identical values, while Naive Bayes had a poorer precision value. In the context of correct classification rates, the Decision Tree model outperformed the other models, showcasing the highest rate, while Naive Bayes had the highest error rate.\\
Kappa statistic values further highlighted the superiority of the Decision Tree model compared to other classification algorithms. For the Holdout method, the Decision Tree model yielded an impressive 99.8\% accuracy. Considering that our pollutant data is of a synthetic type, the Decision Tree model demonstrated its capability to handle this imbalanced dataset effectively, leading to better overall results compared to other models.

In Table \ref{sample}, the performance of the Decision Tree model is evaluated in terms of accuracy, precision, recall, F1-score, Kappa statistic, Root Mean Squared Error, and Matthews Correlation Coefficient with increasing sample sizes. We utilize a data set consisting of 10,559 instances for both training and testing for Decision Tree model. The findings demonstrate that increasing the sample size leads to improved overall performance, with a sample size of 8500 emerging as the preferable choice, exhibiting consistent out-performance across all performance metrics.

In conclusion, the Decision Tree model stood out as the top performer in terms of F1 score, Matthews correlation, precision, correct classification rate, and Kappa statistic value. It exhibited favorable results due to its handling of synthetic imbalanced data, the decision tree model emerged as the most accurate and reliable choice for our specific indoor air quality sensing and estimation task.




 \subsection{Data Interpretation : LIME and SHAP Models}\label{inter}
 Data interpretation tells the process how the predictions are derived from the data using machine learning models. For data interpretation, we are using here LIME and SHAP models.


\begin{figure}
    \centering
    \includegraphics[width=12cm]{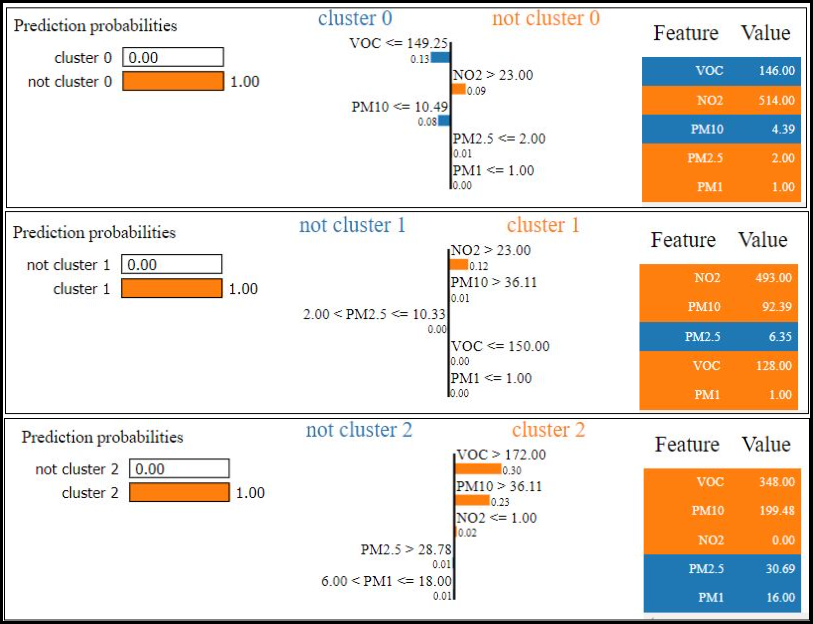}
    \caption{LIME model outcomes presenting the distinctive characteristics of Cluster 0, Cluster 1, and Cluster 2 in three separate subplots}\label{lime}
\end{figure}

\begin{enumerate}
    \item  \textbf{{Outcome of LIME Model:}}
In the LIME (Local Interpretable Model-agnostic Explanations model), the significance of the clusters  lies in their interpretability and local approximation, . LIME aims to provide interpretable explanations for the predictions of complex models by approximating them with simpler, locally-linear models. By understanding the dominant features and their correlations within each cluster, we gain insights into the factors influencing indoor air quality for specific scenarios.
In Figure \ref{lime}, we use the LIME model to explain the changes in the dominant features (NO$_{2}$, VOC, PM$_{10}$, PM$_{1}$) within each cluster that impact in the indoor air quality. Additionally , we can observe how changes in the negatively correlated features (PM$_{2.5}$ and PM$_{1}$) within cluster 2 affect the prediction and, similarly, how changes in PM$_{2.5}$ affect the prognosis in cluster 1.\\
 In cluster 0, we observe that NO$_{2}$, PM$_{2.5}$ is dominant feature among all, and VOC and PM$_{10}$ are negatively correlated.
This level of interpretability allows us to identify critical features that significantly influence indoor air quality in different contexts. It helps stakeholders take appropriate actions to improve air quality based on the specific characteristics of each cluster. For example, in cluster 2, reducing NO$_{2}$, VOC, and PM$_{10}$ levels could be prioritized, while in cluster 1, managing NO$_{2}$, VOC, PM$_{10}$, and PM$_{1}$ becomes crucial.
In cluster 1, the dominant Features: NO$_{2}$, VOC, PM$_{10}$, PM$_{1}$
Where coefficients are, NO$_{2}$ $>$ 23, VOC $\leq$ 150, PM$_{10}$ $>$ 36, and 
 2 $<$ PM$_{2.5}$ $\leq$ 10.33, PM$_{1}$ $\leq$ 1.
As per the Figure \ref{lime} , cluster 2 have NO$_{2}$, VOC, PM$_{10}$
as dominant features.
By leveraging the insights from the LIME model and its clustering, we can make informed decisions to optimize indoor activities and create healthier spaces for occupants.

\item  \textbf{{Outcome of SHAP Model:}}
The SHAP (SHapley Additive exPlanations) model provides valuable insights into how each feature contributes to the overall model output for our prediction. A larger positive SHAP value indicates that the corresponding feature has a higher positive impact on the forecast, while a larger negative SHAP value indicates a higher negative impact.
SHAP model's coefficient revealing the feature importance of each cluster. As per the results, VOC  emerged as the most influential feature, with a SHAP value greater than 0.025. This value indicates a strong positive influence on the model's output, suggesting that higher VOC levels in the indoor environment, emitted from various sources like cooking, smoking, and burning incense sticks, are associated with poorer air quality predictions. 
Next second most, NO$_{2}$ follows closely with a SHAP value of about 0.016, displaying its importance in contributing to the model's output, which is produced mainly from combustion processes, such as cooking and smoking.
Next, PM$_{10}$ is the influential feature, with a SHAP value between the range of 0.010 and 0.015. While it is effective, its impact is slightly less than NO$_{2}$. PM$_{10}$ is commonly generated from cooking, burning paper, and air conditioning, and its presence can lead to respiratory issues and worsen indoor air quality.\\
On the other hand, PM$_{2.5}$ and PM$_{1}$ have lower SHAP values near 0.01, indicating that their contributions are relatively poor compared to other features. \\
In summary, the SHAP model highlights VOC, PM$_{10}$, and NO$_{2}$ as the most important features, meanwhile, the relatively lower SHAP values for PM$_{2.5}$ and PM$_{1}$ suggest that addressing them may still be important, but their individual contributions might have a comparatively smaller impact on the overall air quality prediction.Understanding these feature contributions can aid in interpreting the model's behavior and making informed decisions in the context of the underlying data.



\item  \textbf{{Comparison of LIME and SHAP Models:}}
Based on the provided results and explanations, we can make a comparison between the LIME and SHAP models for interpreting the indoor activity:

1. \textbf{Interpretability approach:}

~~~~~~~~~~   - \textbf{\textit{LIME:}} LIME identifies dominant features and their correlations within each cluster, offering insights into factors influencing indoor air quality in specific scenarios. In our 3 clusters, we achieve clusters dominating features, like cluster 0 has  NO$_{2}$, cluster 1 and cluster 2 has NO$_{2}$ and PM$_{10}$, so each group contains situation-wise pollutants, so we gain knowledge as per the ground truth.

~~~~~~~~~~   - \textbf{\textit{SHAP:}} SHAP values reveal the most influential features for the overall model output. According to the SHAP model's results, VOC, NO$_{2}$, and PM$_{10}$ are the most essential features, with VOC having the highest positive impact on the model's predictions.

2. \textbf{Feature contribution:}

~~~~~~~~~~   - \textbf{\textit{LIME:}} LIME focuses on explaining changes in dominant features (NO$_{2}$, VOC, PM$_{10}$, PM$_{1}$) within each cluster that impact air quality. It allows for understanding how changes in negatively correlated features (PM$_{2.5}$ and PM$_{1}$) in specific clusters affect predictions.

~~~~~~~~~~   - \textbf{\textit{SHAP:}} The SHAP model quantifies the impact of each feature on the model's output. According to the SHAP values, VOC, NO$_{2}$, and PM$_{10}$ have the most significant contributions to the model's predictions, while PM$_{2.5}$ and PM$_{1}$ have relatively lower contributions.\\
Both LIME and SHAP models are utilized for model interpretation. LIME works by fitting a simple local model around a specific prediction, creating a localized explanation. On the other hand, SHAP values represent the contribution of each feature to the prediction. A notable difference between the two approaches lies in their data requirements for generating explanations. LIME only needs one observation to generate an explanation for a prediction. In contrast, the SHAP model requires an entire sample to explain a single value, making it more computationally intensive.
\end{enumerate}

The combined of LIME and SHAP models enables to assign weight factors to different indoor air pollutants based on their importance in degrading indoor air quality. The weight factors determined using a dataset, and the pollutants are ranked. VOCs received the highest weight, followed by PM$_{10}$ and NO$_{2}$, while PM$_{2.5}$ was considered equally harmful to human health. As PM$_{1}$ was found to be less relevant to the study, it was assigned the lowest weight. By considering these weight factors, which aims to provide personalized pollution intake assessments to help individuals understand their potential health risks associated with specific indoor air pollutants in their respective regions.


\section{Personalize indoor air pollution exposure calculation}\label{2}
After assessing the LIME and SHAP models, we determined weighted factors for pollutants, ranking VOC as most impactful, followed by $NO_{2}$ and $PM_{10}$. Using Algorithm 1, we calculated cluster potencies: $C_0$ at 5, $C_1$ at 2, and $C_2$ at 3.6. See Table \ref{rank} for detailed results.
We proceed by summing the main three pollutants for each cluster and then calculating their means. In cluster $C_0$, NO2 dominates with an instance count of 1979.392, while in $C_1$ VOC prevails at 784.29, and in $C_2$ VOC also holds the highest instance count at 1404.75. The ratio of these highest pollutant instances is 5:2:3.6, reflecting their relative importance across the clusters.
Next, we gather individual indoor pollutant data and perform clustering on the targeted person's whole day (24hours) dataset, enabling the calculation of each person's pollution intake over the course of a day.

\begin{table}
\caption{Ranking of clusters in our dataset based on the evaluation results and pollutants' weight factors}
\label{rank}
\begin{tabular} { | c | c | c | c | c | c | c | }
    \hline
Cluster C$_{i}$,    & Pollutant (P$_{j}$) & Sum of pollutant(P$_{j}$)  & Mean of pollutant (P$_{j}$) & Cluster \\
               &                & = Sum($C_i$($P_{jk}$)) &   = Mean($C_i$(P$_{jk}$))  & potency ($x$) \\
    \hline
    \multirow {3} {4em}  & NO$_{2}$(P$_{j}$), j=1 & 494848  & 1979.392 &  \\
& & & & \\
         {C$_{0}$, i =0}     &    VOC(P$_{j}$), j=2 & 174440 & 697.76 &  {5}\\
         & & & & \\
                         & PM$_{10}$   (P$_{j}$), j=3 & 37421.25 & 149.685 &  \\
    \hline
    \multirow {3} {4em}  & NO$_{2}$(P$_{j}$), j=1 & 521191.24  & 53.9089 & \\
    & & & & \\
                       {C$_{1}$, i =1}     &  VOC(P$_{j}$), j=2 & 7582515.72 & 784.29 &  {2}\\
                       & & & & \\
                            & PM$_{10}$   (P$_{j}$), j=3 & 710810.696 & 73.522 &  \\
    \hline
     \multirow {3} {4em}  & NO$_{2}$(P$_{j}$), j=1 & 32096  & 50.15 & \\
     & & & & \\
                       {C$_{2}$, i =2}     & VOC(P$_{j}$), j=2 & 899040 & 1404.75 &  {3.6}\\
                       & & & & \\
                            & PM$_{10}$   (P$_{j}$), j=3 & 277299.2 & 433.28 &  \\
    \hline

    \end{tabular}
\end{table}

In order to determine personalized pollutant intake, we collected data from different habituated individual persons. In Table \ref{exposure}, here we only represent three persons' pollution exposure.
The first person's activities included 24 hours of cooking, using incense sticks, and being in an air-conditioned environment. However, only data from 28th July, 2023, is analyzed after pre-processing the data of 25th July, 26th July, 27th July, and 28th July 2023.
The second person's daily routine consisted of using incense sticks for worship purposes and smoking in an air-conditioned room. Despite having data from 31st July, 1st August, 2nd August, and 3rd August 2023, we used the data from August 3rd, covering the entire 24-hour period, for analysis.
The third person, his habits involved occasional smoking, cooking, and engaging in burning activities. We had data from 4th July to 6th July , 2023, but only the data from 6th July is utilized.

\begin{table}[h!]
\caption{Pollution exposure of 24 hours of our targeted three person}
\label{exposure}
\begin{tabular}{|l|l|l|l|lll|}
\hline
\multicolumn{1}{|c|}{\multirow{2}{*}{\textbf{}}} & \multicolumn{1}{c|}{\multirow{2}{*}{\textbf{Date}}} & \multicolumn{1}{c|}{\multirow{2}{*}{\textbf{Activity}}}                                                                                    & \multicolumn{1}{c|}{\multirow{2}{*}{\textbf{\begin{tabular}[c]{@{}c@{}}Total \\ exposure \\ \end{tabular}}}}      \\ 
\multicolumn{1}{|c|}{}                           & \multicolumn{1}{c|}{}                               & \multicolumn{1}{c|}{}                                                                                                                      & \multicolumn{1}{c|}{}                                                                                                 \\ \hline
Person 1                                         & 25/07/2023                                          & \begin{tabular}[c]{@{}l@{}}Morning - \\ Using incensing \\ stick, Cooking\\ Afternoon - Use\\ of AC\\ 3. Night - Cooking\end{tabular}      & 3531.8                 \\ \hline
Person 2                                         & 01/08/2023                                          & \begin{tabular}[c]{@{}l@{}}Morning - Smoking\\ Afternoon - Use of \\ AC\\ Evening - Use of \\ incense stick\\ Night - Smoking\end{tabular} & 1838                                                                                                                   \\ \hline
Person 3                                         & 04/08/2023                                          & \begin{tabular}[c]{@{}l@{}}Morning - Cooking\\ Afternoon - Smoking\\ Evening - Burning\\ Night - Use of AC\end{tabular}                    & 1676                              \\ \hline
\end{tabular}
\end{table}

By assessing the pollution exposure (as shown in Figure \ref{indoor}) of an individual person, we can employ the equation provided below to determine their total pollutant intake over the course of a day based on their indoor activities.
After clusters the person-wise indoor air pollutant data with k=3.
Now, from the new clustering result we got $C'_{0}$, $C'_{1}$ and $C'_{2}$.\\
Where,
Total $x$ instances(pollutants) belong to $C'_{0}$,\\
$y$ instances (pollutants) belong to $C'_{1}$\\
$z$ instances (pollutants) belong to $C'_{2}$\\
So, the Total exposure = $x$*(cluster potency of $C_{0}$) + $y$*(cluster potency of $C_{1}$) + $z$*(cluster potency of $C_{2}$).
The accurate result is shown in Table \ref{exposure}. Person 1 has a higher total exposure than both Person 2 and Person 3.

\begin{table}[]
\centering
\caption{Pollutant wise pollution exposure after the normalization }
\label{exposure1}
\begin{tabular}{|l|l|l|l|}\hline
         & VOC  & $NO_{2}$  & $PM_{10}$  \\\hline
Person 1 & 1.21 & 3.52 & 15.8  \\\hline
Person 2 & 1.02 & 1.35 & 1.00  \\\hline
Person 3 & 1.00 & 1.00 & 6.70 \\\hline
\end{tabular}
\end{table}

\begin{sidewaysfigure}  \includegraphics[width=\paperwidth]{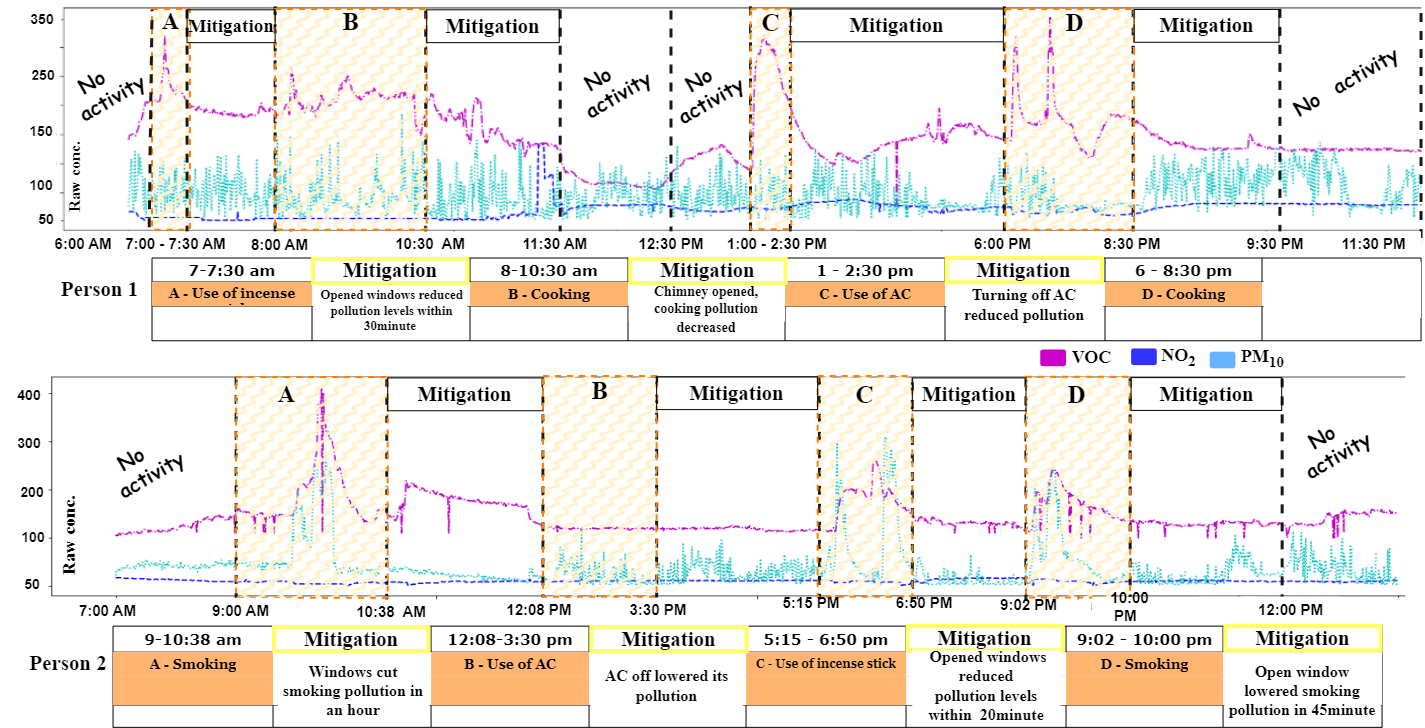}
\caption{Prediction of 24-hour indoor air pollution activities, with A representing the first activity, B as the second, C as the third, and D as the fourth activity of an individual.}\label{activity}
\end{sidewaysfigure}

The result (Table \ref{exposure}) shows that  Person 1 has a higher total exposure than both Person 2 and Person 3. Person 1 has the highest total exposure due to engaging in cooking twice a day, burning incensing stick, and using the AC. Person 2 smokes twice a day and uses incense sticks and the AC, while Person 3 uses the AC, smokes, burns, and cooks once in 24 hours, resulting in lower pollutant intake. Cooking and smoking both contribute to VOC emissions. However, Person 1's extended cooking time (approximately 5 hours daily, (morning 8-10:30 am and evening 6 to 8:30 pm)) leads to a higher VOC intake of 1.21, as shown in Table \ref{exposure1}, compared to Person 2's smoking intake of 1.02 (approximately 2.5 hours daily, morning 9-10:38 am and evening 9 to 10 pm)). Figure \ref{activity} illustrates the graphical representation of the complete time schedules for person 1 and person 2. Since VOC is the most dominant pollutant, Person 1's pollution intake surpasses the others. Our approach can accurately calculate pollution exposure from various indoor activities, offering personalized recommendations to mitigate health risks. This self-personified pollution measurement model alerts individuals, empowering them to protect themselves from indoor pollution.


This analysis allows for the assessment of health risks and empowers individuals to identify activities as well as pollutants that may pose greater harm to their well-being.

\section{Conclusion and Future scope}\label{sec7}
Our study explores into the critical issue of indoor air pollution, shedding light on its often overlooked but significant health implications. We examined the impacts of various indoor activities, uncovering their contributions to pollution levels and human health risks. The findings from our online survey underscore the alarming lack of awareness about indoor air pollution among respondents.
Our research monitored indoor pollution and identified key contributors, ranging from cooking and smoking to indoor appliances like air coolers and incense sticks. Through data collection, clustering analysis, and interpretability techniques, we established a personalized pollution assessment framework, allowing us to pinpoint specific sources of pollution for individuals. This empowers targeted interventions and informed decision-making, ultimately enhancing indoor air quality and safeguarding individual well-being.
Looking ahead, our research opens avenues for further advancements. 

A future focus could involve the development of a real-time monitoring device that collects a person's daily pollution exposure, diagnoses health effects, and provides timely alerts. For instance, the device could recommend opening windows for fresh air intake in response to excessive AC usage, advise reducing cigarette consumption upon exceeding benchmark levels, or suggest using exhaust fans during high-pollution cooking sessions. Such a personalized alert system could transform individuals' behavior toward mitigating pollution-related health risks. By bridging the gap between awareness and action, our research lays the foundation for a proactive approach to indoor air quality management, facilitating healthier indoor environments and minimizing the health hazards associated with indoor air pollution.


    
\bibliography{sn-bibliography}

\section*{Declarations}
\begin{itemize}
\item Funding - This research received no specific grant from any funding agency.
\item Authors Contributions - Pritisha Sarkar - original manuscript writing, methodology data acquisition; Kushalava reddy Jala - Software; Mousumi Saha - methodology, supervision.
\item Competing interests - The authors affirm that there are no identifiable conflicting financial interests or personal affiliations that could have potentially influenced the findings presented in this paper.
\item Ethics approval - does not require ethics approval
\item Consent to participate - Not applicable
\item Consent for publication - Not applicable
\item Availability of data and materials - The dataset will be provided upon reasonable request.
\item Data availability - Data will be provided upon reasonable request.
\item Code availability - Not applicable
\end{itemize}

\end{document}